\documentclass{article}

\usepackage{iclr2027_conference,times}

\usepackage{amsmath,amsfonts,bm}

\def\eqref#1{equation~\ref{#1}}

\def\1{\bm{1}}

\DeclareMathAlphabet{\mathsfit}{\encodingdefault}{\sfdefault}{m}{sl}
\SetMathAlphabet{\mathsfit}{bold}{\encodingdefault}{\sfdefault}{bx}{n}

\usepackage{amsmath,amssymb,mathtools}
\usepackage{booktabs}
\usepackage{graphicx}
\usepackage{microtype}
\usepackage[hidelinks]{hyperref}
\usepackage{url}

\title{Do world models learn global understanding?}

\author{Alexander Detkov \& Matt Thomson \\
California Institute of Technology \\
\texttt{\{adetkov,matt\}@caltech.edu}}

\iclrfinalcopy

\begin{document}

\maketitle
\lhead{Preprint. Under review.}

\begin{abstract}

AI systems often feel frustratingly brittle and fragmented. A large language model (LLM) may correctly explain a concept but fail to apply it, or follow safety instructions in one context but not another. This suggests a failure to lift local information to global understanding. To gain fundamental insight, we frame ``understanding'' as learning constraints and propagating their consequences. We construct learning tasks on monoid worlds, sets of states connected by action transitions, where observed transitions and an unseen constraint jointly determine held-out transitions, so generalization tests whether models infer global constraints from local transitions and propagate them. Across attention, recurrent, and state-space architectures, next-state training on paths of observed transitions fits the data but fails to propagate non-trivial constraints. Compositional training, which uses identical paths but hides intermediate states from the input, propagates inverse, commutative, and composition constraints at \(96\%\) accuracy against under \(1\%\) for next-state training. A linear analysis shows how the compositional objective incentivizes a shared state representation that propagates constraints. Improvements hold at scale and in realistic domains: 174M-parameter world models of scanned homes propagate geometric constraints at \(66\%\) against \(13\%\), and 4B--14B LLMs finetuned on naturalistic articles propagate composition constraints at \(81\%\) against \(0\%\). World models used for control through search, imagination, or representation reach goals optimally through held-out transitions \(78\%\) of the time when compositionally trained, against \(13\%\) when next-state trained. Finally, to measure how far constraints propagate, we introduce proof depth, the minimum number of inference rounds needed to derive a held-out transition. Accuracy falls from \(95\%\) at depth one to \(49\%\) at depth two, but training on longer paths improves deeper propagation. These results provide a formal way to investigate global understanding and show that compositional training promotes information propagation and integration.
\end{abstract}

\section{Introduction}
What does it mean for a system to understand? Do language models understand semantics when training on ``Tom's parent is Mary'' fails to teach ``Mary's child is Tom'' \citep{berglund2024reversal}? Do world models understand space if a left rotation followed by a right rotation produces a new scene \citep{kwon2026stable}? In both cases, a constraint (parent and child are converses; left and right rotations are inverses) lets a few seen facts determine many held-out ones. This suggests framing ``understanding'' as learning constraints and propagating their consequences globally.

To study constraint learning and propagation systematically, we construct monoid worlds where constraints are controllable and their consequences mathematically determined. A monoid world is a set of states with actions defining transitions between them. In space, states could be points on a line and actions \(\text{left}/\text{right}\) translations. These actions satisfy an inverse constraint: moving left then right returns to the original state. Analogously, semantic relations obey constraints like composition: the parent of a parent is a grandparent. We capture these constraints as equalities over action sequences, letting us determine from data, manipulate, and directly specify each world's constraints.

Since each world is precisely defined, we can directly evaluate whether models propagate constraints by constructing held-out sets of transitions which are uniquely determined (semantically entailed) by joint knowledge of the constraint and observed transitions. For example, observing \((2, \text{left}, 1)\) entails the held-out transition \((1, \text{right}, 2)\) under the constraint \(\text{left}=\text{right}^{-1}\). Generalization to held-out transitions tests whether models learn constraints from data and propagate their consequences.

Not all consequences of constraints are created equal, and some require multiple inference steps, as when a new fact sets off a chain of updates. We frame this formally by defining a proof system and measure the inference complexity of a held-out transition by its proof depth \(d\): the minimum number of rounds to derive the transition. We construct worlds whose held-out transitions vary in proof depth and explore how generalization and training dynamics change with the required inference depth.

Next-state training on paths of monoid transitions fails to propagate non-trivial constraints across transformer, recurrent, and state-space models. Compositional training, which learns action compositions from the same paths by hiding intermediate state inputs, averages \(96\%\) on inverse, commutative, and composition constraints, and a linear analysis shows how it yields a shared state representation that propagates constraints. We apply the monoid framing to world models of visual environments and to pretrained LLMs learning relational structure. Traditional next-state-style objectives poorly propagate geometric and semantic constraints, while compositional training improves propagation of commutative constraints in world models by \(+42\%\) and of composition constraints in LLMs by \(+73\%\), with gains that hold in 174M-parameter world models of scanned homes and in 4B--14B LLMs finetuned on naturalistic articles. Used for control through search, imagination, or representation, compositionally trained world models reach goals optimally through held-out transitions \(78\%\) of the time, against \(13\%\) for next-state models. Finally, generalization decreases sharply with proof depth \(d\) and transitions are learned in order of depth, suggesting that models apply constraints iteratively but with difficulty, while increasing composition length \(T\) improves deeper propagation.

\section{A monoid framework for global constraints}
We construct worlds governed by specified constraints and hold out transitions which are jointly implied by the remaining seen transitions and the constraints. We train models on paths of seen transitions and evaluate them on held-out transitions, measuring a model's ability to learn global constraints from local data and propagate their consequences.

\subsection{Monoid world and global constraints}
\begin{figure*}[h]
    \centering
    \includegraphics[width=\textwidth]{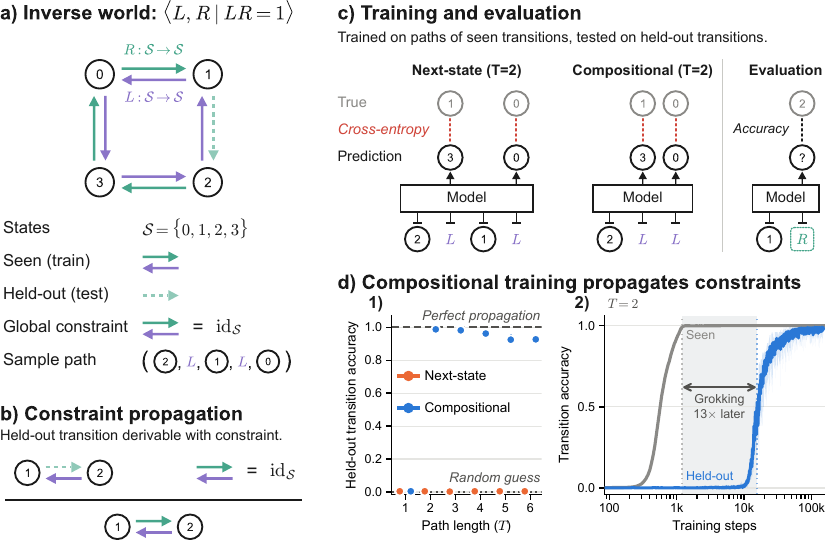}
    \caption{\textbf{An inverse monoid world with training and evaluation setup.} \textbf{a)} A four-state world with actions \(L\) and \(R\), constraint \(LR=RL=1\) and sample training path. \textbf{b)} A seen transition and the constraint derive the held-out transition. \textbf{c)} Next-state training shows intermediate states, while compositional training does not. Evaluation is generalization over held-out transitions. \textbf{d)} \(N=1024\) worlds, five seeds. 1) Next-state training fails to propagate inverse constraint, while compositional training succeeds (\(\geq93\%\)). 2) Training curves for \(T=2\) show sharp grokking-like generalization.}
    \label{fig:inverse_world}
\end{figure*}

A monoid world is a finite set of states \(\mathcal S:=[N]:=\{0, \dots, N-1\}\) with a set of actions \(\mathcal A\), each a function \(a: \mathcal S\to\mathcal S\) on the state space. Consider the world shown in Figure \ref{fig:inverse_world}a, which has a discrete \(90^\circ\) rotation interpretation with states \(\mathcal S:=[4]\) and actions \(L: x\mapsto x-1 \bmod 4\) and \(R: x\mapsto x+1 \bmod 4\). Actions can be composed to form action sequences, and with the identity action \(1\), form a monoid. Local observations are transitions \((s, a, a(s))\), which correspond to the change of state \(s \in \mathcal S\) under the action \(a\in\mathcal A\). In our rotation world, \((0, R, 1)\) is a transition.

The global constraints of our monoid world are equality relations between action sequences: two distinct sequences of actions induce the same map on the entire state space. In other words, at every initial state, the paths of transitions for the two action sequences converge to the same terminal state. The global nature comes from the requirement that the constraint on transitions applies uniformly over the entire state space. In our rotation world, we have two essential constraints: inverse \(LR=RL=1\) and periodicity \(L^4=1\), from which all other constraints are derived.

When building a monoid world with a specified constraint, we choose a state-space size \(N\) and randomly sample each action's map such that the constraint is satisfied. For example, when constructing inverse worlds \(LR=RL=1\) as in Figure \ref{fig:inverse_world}a, we choose \(L: \mathcal S \to \mathcal S\) as a random bijection and derive \(R\) as its inverse. Monoid theory warns us that finite worlds may inadvertently introduce unwanted constraints, so we audit all our worlds by enumerating \(u=v\) global constraints for \(|u|+|v|\leq 8\) and reject impure worlds. We explore inverse, commutativity, composition, periodicity, and copy constraints shown as a presentation, \(\langle \text{Actions} \mid \text{Constraints}\rangle\), in Figure \ref{fig:arch_relations}.

\subsection{Evaluation of constraint propagation by entailed hold-outs}

We hold out transitions which are jointly inferable from observed training transitions and the world's global constraint. Formally, inferability is captured by semantic entailment.  A transition \((s, a, s')\) is semantically entailed by constraint \(\mathcal C\) and seen transitions \(\mathcal T\) if and only if every choice of action functions on \(\mathcal S\) that satisfies \(\mathcal C\) and agrees with \(\mathcal T\) also contains it. Consider the cyclic inverse world in Figure \ref{fig:inverse_world}a. The held-out transition \((1, R, 2)\) is semantically entailed since any world that contains the seen transition \((2, L, 1) \in \mathcal T\) and satisfies the inverse constraint must contain it. In practice, we perform greedy sampling with semantic entailment checks until we reach a specified hold-out size. 

\subsection{Next-state and compositional training}
We train models on paths of \(T\) transitions, \((s_0,a_0,s_1,\ldots,a_{T-1},s_T)\), where \(s_{t+1}=a_t(s_t)\). The training set is every such path made only of seen transitions, sampled uniformly. Models never receive the global constraint directly. Both objectives predict \(s_t\) for \(1\leq t\leq T\) and differ only in their inputs (Figure \ref{fig:inverse_world}c). Next-state training shows intermediate states, \((s_0,a_0,s_1,\ldots,s_{t-1},a_{t-1})\), while compositional training hides them, \((s_0,a_0,a_1,\ldots,a_{t-1})\), forcing the model to compose actions internally, with \(T\) setting the composition length. The two coincide at \(T=1\). We evaluate top-1 accuracy on held-out one-step queries \((s,a)\mapsto a(s)\). No evaluation requires composing actions.

\subsection{Measuring inference complexity of a transition with proof depth}
Certain held-out transitions are intuitively harder to infer than others. In the commutative world of Figure \ref{fig:commutative_world}, both held-out transitions are semantically entailed, but only one can be immediately derived from seen transitions; the other requires first inferring an intermediate transition. We quantify this dependency by proof depth, defined relative to a fixed set of inference rules based on the world's algebraic constraints. Consider the rules of the commutative world depicted in Figure \ref{fig:commutative_world}b. Knowing transitions \((1, F, 0), (0, R, 2)\), and \((1, R, 3)\) derives \((3, F, 2)\) since actions \(F\) and \(R\) commute, so the paths \(1 \xrightarrow{F} 0 \xrightarrow{R}2\) and \(1 \xrightarrow{R} 3 \xrightarrow{F}2\) must end at the same state. A constraint cascade proceeds in rounds: each round applies the inference rules to the seen transitions and to those derived in earlier rounds. The proof depth of a transition is the first round in which it is derived.

\section{Constraint propagation in controlled monoid worlds}
\label{sec:monoid_world}

We construct monoid worlds with \(N=1024\) states satisfying specified constraints and hold out \(10\%\) of one action's transitions \((s, a, a(s))\), each entailed by seen transitions and the constraint. Models train on paths of seen transitions and are evaluated on held-out transitions. Unless stated otherwise, models have four layers of width 64 and train for \(150\)k steps at \(T=2\), averaged over five seeds.

\subsection{Next-state training fails to propagate constraints}
Under a next-state objective, where intermediate states are shown, transformers on inverse worlds fail to generalize with accuracy at chance (Figure \ref{fig:inverse_world}d.1). Scaling model, world, and compute does not save next-state training, and models continue to fail at propagating inverse constraints (Appendix \ref{appx:scale_next_state}).
  
Across attention, recurrent, and state-space architectures, next-state training at \(T=2\) fails, indistinguishable from chance, at propagating inverse, commutative, composition, and periodicity constraints in their associated worlds (Figure \ref{fig:arch_relations}). Only in the trivial copy world, where the two actions are identical (\(A=B\)), does next-state training generalize.

We find a systematic ``copy'' failure mode of general constraint propagation by next-state trained models, where models behave as if every world were the copy world. Averaged across architectures and worlds, \(64\%\) of incorrectly inferred held-out transitions are explained by copy propagation, assuming the held-out \((s, A, A(s))\) transition matches seen \((s, B, B(s))\), although seen transitions contradict the \(A=B\) constraint at almost every state (Appendix \ref{appx:copy_bias_next_state}). These results suggest the next-state objective poorly propagates constraints from data.

\subsection{Compositional training improves constraint propagation}
\begin{figure*}[t]
    \centering
    \includegraphics[width=\textwidth]{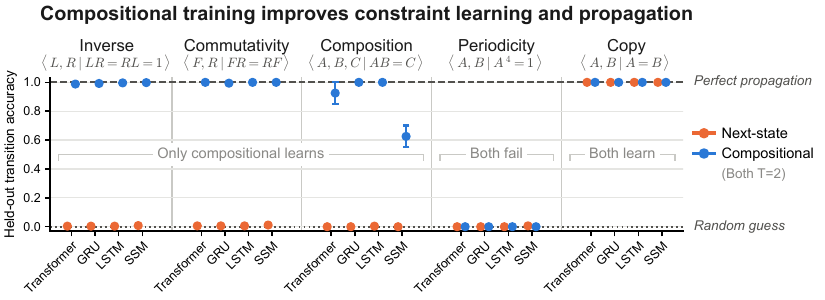}
    \caption{\textbf{Across monoid worlds and architectures, compositional training propagates constraints that next-state training does not, with equal optimizer steps.} Compositional training averages \(96\%\) held-out accuracy on inverse, commutative, and composition constraints. Next-state training succeeds only on copy, and neither propagates periodicity. Mean \(\pm\) SD over five seeds.}
    \label{fig:arch_relations}
\end{figure*}

Under a compositional objective, where intermediate states are hidden, transformers successfully propagate inverse constraints, generalizing held-out transitions with \(\geq 93\%\) accuracy (Figure \ref{fig:inverse_world}d.1). We find a sharp jump in generalization of \(+98\%\) when varying the composition length from \(T=1\), a reduction to next-state training, to \(T=2\), when training first includes two-action compositions.

Across architectures, compositional training at \(T=2\) propagates inverse, commutative, and composition constraints at \(96\%\) on average, and copy at \(100\%\) (Figure \ref{fig:arch_relations}). Extending composition length to \(T=4\), accuracy is the same  \(96\%\) (Appendix \ref{appx:composition_length}). Compositionally trained models also have reduced ``copy'' failure modes, hallucinating a copy constraint for \(36\%\) of failures (Appendix \ref{appx:copy_bias_next_state}). Neither objective propagates periodicity constraints at \(T=2\) or \(T=4\). Across worlds \(A^k=1\) for \(k=3,4,5\), compositionally trained transformers fail at every path length up to \(T=8\), even when paths contain the whole constraint, and longer training does not help. Appendix \ref{appx:periodicity} traces the failure to periodicity being the only constraint that involves a single action.

Compositional training fits seen transitions an order of magnitude earlier than it recovers held-out ones (Figure \ref{fig:inverse_world}d.2). On the inverse world, transformers fit \(100\%\) of seen transitions by \(1.2\)k steps but reach \(50\%\) held-out accuracy only at \(16\)k, about \(13\) times as many steps, a grokking-like delay \citep{power2022grokking}. Finally, at fixed world size, held-out accuracy increases monotonically with ``constraint evidence'', the number of seen transitions satisfying the constraint (Appendix \ref{appx:evidence_scaling_compositional}).

\subsection{Why compositional training generalizes and next-state training does not}
\label{sec:shared_space}
A held-out transition \((s, R, t)\) follows from its seen reverse \((t, L, s)\) only through the constraint \(LR=1\). A linear model, which captures how our nonlinear models generalize (Appendix~\ref{appx:linear_evidence}), illustrates how compositional training can recover the constraint. Each state \(s\) is a vector \(x_s\), each action a matrix \(M_a\), and the next state is read out from \(U M_a x_s\). Compositional training fits \(U M_b M_a x_s\) and \(U M_b x_{a(s)}\) to the same target, which can encourage \(M_a x_s\) to align with \(x_{a(s)}\). In the idealized case of a shared state representation, where \(M_a x_s = x_{a(s)}\) on seen transitions, each seen path out by \(L\) and back by \(R\) gives \(M_R M_L x_u = x_u\) at its starting state \(u\). If these vectors span the embedding space, as they generically do when starting states outnumber embedding dimensions, then \(M_R M_L = I\). The held-out transition follows: \(x_s = M_L x_t\), so \(M_R x_s = M_R M_L x_t = x_t\), giving the same readout as on seen transitions into \(t\). Next-state training fits only one-step predictions, unchanged when every \(M_a\) is replaced by \(G M_a\) and \(U\) by \(U G^{-1}\) for any invertible \(G\). Its loss thus forces neither \(M_a x_s = x_{a(s)}\) nor \(M_R M_L = I\), and nothing ties \(x_s\) to \(M_L x_t\). Trained linear models approximate these equations much more closely under compositional training (Appendix~\ref{appx:shared_representation}).

\section{Constraint propagation in vision and language}
\begin{figure*}[t]
    \centering
    \includegraphics[width=\textwidth]{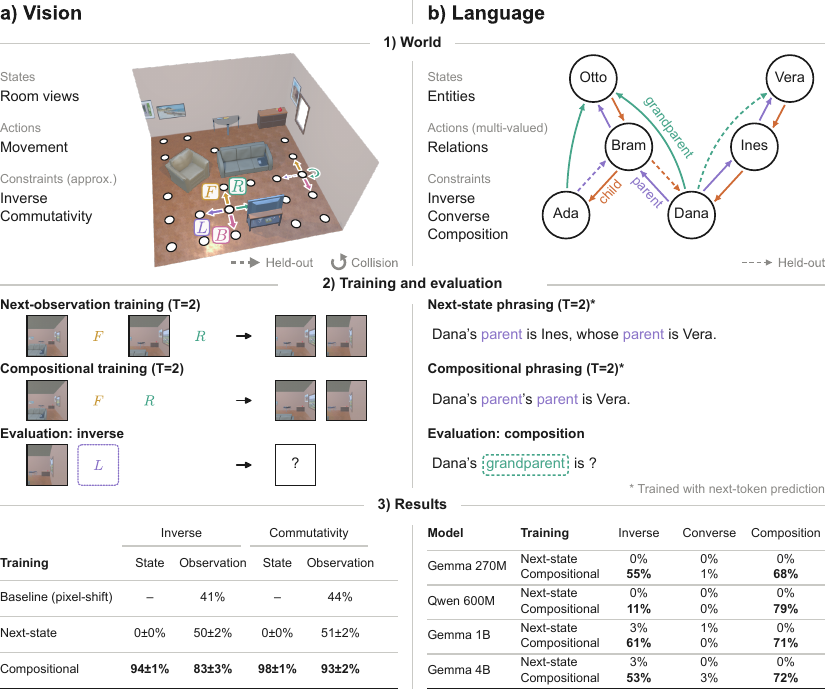}
    \caption{\textbf{Compositional objectives improve constraint propagation over next-state-style training in vision and language.} \textbf{a)} 1) An agent moves through an AI2-THOR room, where inverse \(LR=RL=1\) and commutative \(FR=RF\) constraints hold except where collisions block a move. 2) Training follows Figure \ref{fig:inverse_world}c with observations in place of states. 3) From observations, compositional training beats next-state training by \(+38\%\) and a pixel-shift baseline by \(+46\%\). From states, compositional training reaches \(96\%\) and next-state training stays at chance. Mean \(\pm\) SD over three seeds. \textbf{b)} 1) A Wikidata knowledge graph with entities as states and relations as multi-valued actions; the inverse pair, successor/predecessor, is not drawn. 2) LLMs are finetuned with next-token prediction on either phrasing. 3) Compositional phrasing propagates inverse (\(45\%\)) and composition (\(73\%\)) but not converse. Next-state phrasing propagates none. Averages over four LLMs, one seed.}
    \label{fig:vision_language}
\end{figure*}

We test constraint propagation in world models trained on visual environments and in LLMs finetuned on knowledge-graph facts. In each domain we identify constraints, which often hold only approximately, and build entailed hold-outs to test their propagation.

\subsection{Geometric constraints in visual world models}
\label{sec:vision_world}

An agent moves (L)eft/(R)ight/(F)orward/(B)ackward through an AI2-THOR virtual environment \citep{kolve2022ai2thor}, generating observation-action paths of length \(T=2\) (Figure \ref{fig:vision_language}a.1). Euclidean geometry constrains motion in two ways. The endpoint does not depend on the order of translations: stepping forward then right, or right then forward, reaches the same place. And every step can be undone by the opposite step. Object collisions break these constraints at some states, so we measure approximate constraints from data, declaring one if two action sequences reach the same terminal observation from at least \(50\%\) of initial observations. Among those found are inverse \(LR=RL=1\) and commutativity \(FR=RF\), which hold at roughly \(80\%\) of states.

We hold out transitions \((\text{obs}, \text{action}, \text{next obs})\) that test the inverse and commutative constraints at states away from collisions, where the constraints hold. Transformer world models predict each next observation, and a held-out transition is correct when the true observation is the most likely one in the dataset (Figure \ref{fig:vision_language}a.2). Since visual similarity alone may recover held-out transitions, we compare to a pixel-shift baseline that translates the current observation and returns the closest match.

Under next-observation training, where intermediate observations are shown, world models reach only \(50\%\) on inverse and \(51\%\) on commutative hold-outs, against \(41\%\) and \(44\%\) for the pixel-shift baseline (Figures \ref{fig:vision_language}a.2 and \ref{fig:vision_language}a.3). Trained compositionally, hiding intermediate observations, world models reach \(83\%\) and \(93\%\), an average gain of \(+38\%\) over next-observation training.

To remove visual cues, we replace observations with state IDs in inputs and targets. Next-state training falls to chance (\(0\%\)) on both, while compositional training reaches \(94\%\) on inverse and \(98\%\) on commutative hold-outs (Figure \ref{fig:vision_language}a.3). Next-state-style training propagates geometric constraints poorly in visual domains, while compositional training propagates them even without visual cues.

Propagation improvements hold at scale in environment, model, and actions. We train on 12 scanned real homes from Habitat HM3D \citep{ramakrishnan2021hm3d}, where the agent steps relative to its heading and also turns, adding periodicity and rotation constraints. Compositionally trained 174M-parameter world models, 60 times larger than those in AI2-THOR, propagate constraints at \(66\%\) from observations and \(69\%\) from states, against \(13\%\) and \(6\%\) for next-state training (Appendix \ref{appx:geometric_scale}).

\subsection{Relational constraints in large language models}
\label{subsec:llm}
We examine whether LLMs trained on facts propagate relational constraints without premises supplied in context. We study inverse (successor and predecessor), converse (parent and child), and composition (a parent's parent is a grandparent). Converse differs from inverse because relations branch: the child of a person's parent may be a sibling rather than the person. For each constraint, we extract a Wikidata knowledge graph of \(\sim512\) renamed individuals and their relations (Figure \ref{fig:vision_language}b.1) and hold out relationships entailed by the training relationships and the constraint. For example, \((\text{Dana},\text{parent},\text{Bram})\) entails \((\text{Bram},\text{child},\text{Dana})\) under converse.

We finetune four pretrained base LLMs (Gemma and Qwen, 270M to 4B parameters) with next-token prediction on natural-language sentences generated from \(T=2\) training paths, comparing two phrasings of the same path. Next-state phrasing states intermediate entities, e.g., ``Dana's parent is Ines, whose parent is Vera'', while compositional phrasing omits them, ``Dana's parent's parent is Vera'' (Figure \ref{fig:vision_language}b.2). Unlike compositional training in monoid worlds, next-token prediction over compositional phrasing never uses intermediate entities as targets, so we include \(T=1\) paths in both conditions. We evaluate by prompting a held-out relationship, e.g., ``Bram's child is'', and choosing the highest-likelihood name in the graph, excluding other correct answers (e.g., Dana's siblings).

Next-state phrasing fails to propagate inverse and composition constraints, with an average held-out accuracy of \(1\%\) (Figure \ref{fig:vision_language}b.3). Compositional phrasing propagates inverse at an average of \(45\%\) and composition at \(73\%\). Neither phrasing propagates converse. Propagation improvements hold at scale, with 4B to 14B parameter LLMs trained compositionally on thousands of naturalistic articles averaging \(81\%\) on composition questions against \(0\%\) for next-state training (Appendix \ref{appx:composition_scale}).

Composition over paths offers a shortcut: although ``Dana's grandparent is Vera'' is never shown, models see the \(T=2\) path ``Dana's parent is Ines, whose parent is Vera'' under next-state phrasing, or ``Dana's parent's parent is Vera'' under compositional phrasing. Does propagation survive without such shortcut paths? We train transformers on \(T=2\) paths in the \(N=1024\) monoid composition world with and without shortcuts (Appendix \ref{appx:composition_shortcut}). Next-state training fails in both settings, while compositional training still recovers \(50\%\) without shortcuts, against \(92\%\) with them. These results suggest that while shortcuts help recovery, they are not required for composition propagation.

Why does converse fail? Parents in Wikidata have \(2.5\) children on average, so the converse relation branches. To isolate branching, we build an \(N=1024\) family tree where each parent has one child and a second with probability \(\beta\), ranging from a line (\(\beta=0\)), where parent and child are inverses, to a binary tree (\(\beta=1\)) (Appendix \ref{appx:branching}). A compositionally trained transformer (\(T=2\)) reaches \(99\%\) on the line, \(63\%\) at \(\beta=0.1\), and \(\leq 3\%\) for \(\beta \geq 0.75\). Branching acts as noise on the inverse constraint: parent then child returns to the start only sometimes. Adding random transitions to the line world with the same L2 deviation from the constraint lowers accuracy equally (Figure \ref{fig:branching}), while noise that preserves constraints in probabilistic worlds leaves propagation intact (Appendix \ref{appx:probabilistic}).

Compositional phrasing improves inverse and composition propagation in LLMs. Converse fails because branching acts as noise on the inverse constraint.

\section{Constraint propagation in planning and control}

\begin{figure*}[t]
    \centering
    \includegraphics[width=\textwidth]{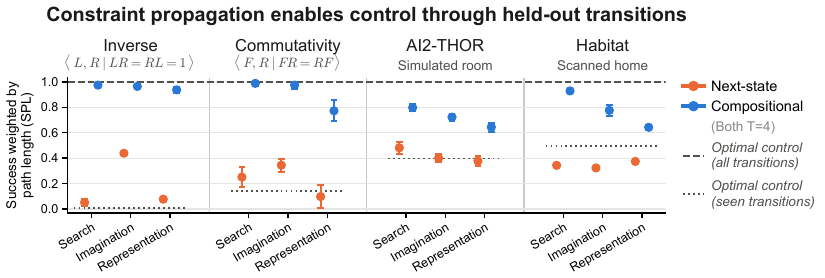}
    \caption{\textbf{Across worlds and control methods, compositionally trained world models reach goals through held-out transitions more efficiently than next-state trained ones.} SPL of goal reaching by search in the world model, a policy learned in its imagination, and a policy learned on its frozen representations, in the inverse and commutative monoid worlds, an AI2-THOR room, and a Habitat home. Both objectives train at \(T=4\). Compositional models beat next-state models with every method in every world, and search with them is near the ceiling. Dashed and dotted lines show optimal control with all transitions and with seen transitions only. Mean \(\pm\) SD over five seeds.}
    \label{fig:control}
\end{figure*}

World models are used for control in three standard ways: search in the model \citep{schrittwieser2020muzero}, policy learning in its imagination \citep{hafner2020dreamer}, and policy learning on its representations \citep{ha2018worldmodels}. We compare next-state and compositionally trained transformer world models (\(T=4\)) under all three. The task is to reach a goal \(g\) from a state \(s\) in the fewest actions, in the inverse and commutative monoid worlds and in randomly explored AI2-THOR and Habitat environments. Test goals lie within four actions, and every shortest path to them crosses a held-out but entailed transition, so reaching them optimally requires inference by the world model.

We report success weighted by path length (SPL) \citep{anderson2018evaluation}: the shortest path length over path length taken if the goal is reached, and \(0\) otherwise. Search finds shortest paths through the model's predicted transitions. Imagination and representation train a policy \(\pi(a \mid s, g)\) on the frozen state embeddings, by reinforcement learning in the model's imagination or behavior cloning on goals with fully seen shortest paths. We compare against optimal control with all transitions (ceiling) and seen transitions only (no propagation). Details, success rates, and optimal rates are in Appendix \ref{appx:control}.

\subsection{Constraint propagation enables control through held-out transitions}
Only world models that propagate constraints enable effective planning and control through held-out transitions. Search in compositional models reaches near-optimal SPL of \(0.97\) and \(0.99\) in the inverse and commutative worlds, against \(0.05\) and \(0.25\) for next-state models (Figure \ref{fig:control}). In AI2-THOR and Habitat, compositional training improves search SPL by \(+0.32\) and \(+0.58\). Policies learned in imagination inherit propagated constraints, improving SPL by \(+0.48\) on average. In the monoid worlds, next-state imagination beats next-state search only by acting randomly: an agent that follows seen transitions and acts randomly otherwise scores similarly (\(0.45\) and \(0.39\), against \(0.44\) and \(0.34\)).

Search and imagination both use the world model's predicted transitions. A next-state world model mispredicts held-out transitions (Appendix \ref{appx:copy_bias_next_state}), so control either detours through seen transitions or, worse, follows transitions that do not exist. With search or imagination, next-state models reach test goals by a shortest path at most \(29\%\) of the time, against \(59\)--\(99\%\) for compositional models (Appendix \ref{appx:control}). In Habitat, next-state search plans through hallucinated transitions and fails on \(40\%\) of test goals, against \(4\%\) for control with seen transitions only, so mispredictions actively harm planning.

\subsection{Representations transfer propagated constraints to learned policies}
Search and imagination use the world model's predicted transitions directly; a policy learned on representations does not. The policy is trained only on goals with fully seen shortest paths, so it can use held-out transitions only if propagated constraints are encoded in the geometry of the state embeddings. Policies on compositional representations reach an average SPL of \(0.75\) across worlds, against \(0.23\) on next-state representations, \(0.26\) for optimal control with seen transitions, and \(0.19\) for a policy on random embeddings (Figure \ref{fig:control}, Appendix \ref{appx:control}). Compositional representations therefore transfer propagated constraints without search or training on held-out transitions, consistent with the shared state representation that compositional training learns (Section~\ref{sec:shared_space}).

\section{How far do constraints propagate?}
\begin{figure*}[t]
    \centering
    \includegraphics[width=\textwidth]{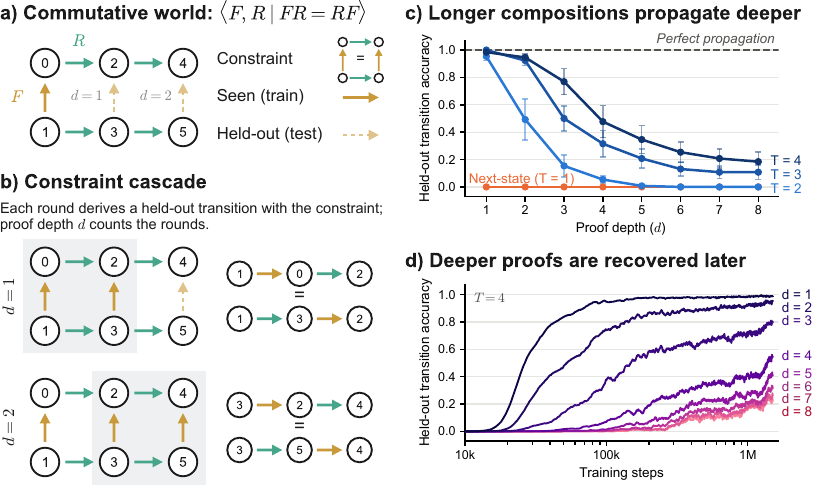}
    \caption{\textbf{Proof depth measures how far a constraint must propagate and predicts generalization and training order.} \textbf{a)} Part of a commutative world with two held-out transitions. \textbf{b)} A constraint cascade applies the constraint in rounds. Proof depth \(d\) is the round that derives a transition, so the \(d=1\) transition is a premise for \(d=2\). \textbf{c)} Accuracy falls with proof depth, and longer composition length \(T\) extends propagation to deeper transitions. Mean \(\pm\) SE over five seeds. \textbf{d)} Transitions are recovered in order of proof depth, suggesting an internal constraint cascade. Mean over five seeds.}
    \label{fig:commutative_world}
\end{figure*}

We measure a held-out transition's derivation complexity by proof depth and construct commutative monoid worlds of size \(N=1024\) with the \(10\%\) held-out transitions spread nearly uniformly across depths \(d=1 \text{ to } 8\) (Figure \ref{fig:commutative_world}a). Previous results concerned only \(d=1\) generalization. We now explore the effect of proof depth on constraint propagation by compositionally trained transformers. We train with composition lengths \(T=1,2,3,4\) (\(T=1\) is next-state training) for \(1.5\)M steps over five seeds. 

\subsection{Constraint propagation decreases sharply with proof depth}
We find generalization of compositionally trained models decreases sharply with proof depth, while next-state trained models do not generalize at any proof depth. A transformer trained with composition length \(T=2\) recovers \(d=1\) transitions, whose derivation depends directly on seen transitions, at \(95\%\), while \(d=2\) recovery, transitions whose derivation also depends on held-out \(d=1\) transitions, is only at \(49\%\). The steep decreasing trend in generalization continues to deeper transitions with \(d=3\) at \(15\%\), \(d=4\) at \(5\%\), and \(d=5\) onward indistinguishable from random chance. Similar step-to-step accuracy ratios over depths \(1\leq d \leq 4\) suggest approximate geometric decay with rate \(0.4\pm0.1\). Even with compositional training, models struggle to propagate constraints deeply.

\subsection{Propagation dynamics suggest internal constraint cascades}
Models trained compositionally propagate constraints to infer transitions iteratively by proof depth: first \(d=1\) transitions are learned, then \(d=2\), \(d=3\), and so on (Figure \ref{fig:commutative_world}d). Transformers trained with composition length \(T=4\) reach half their final accuracy for \(d=1\) at \(26\)k training steps, \(d=2\) at \(50\)k steps, \(d=3\) at \(102\)k, and \(d=4\) at \(236\)k, so each depth takes about twice as many steps as the one before (\(2.1\times\), geometric mean). Monotonic ordering holds for all \(1\leq d \leq 8\), but spacing beyond \(d=4\) shrinks, and \(d=5\) to \(d=8\) all reach half their final accuracy between \(364\)k and \(376\)k steps.

Timing suggests that models may be propagating constraints in a qualitatively similar way to constraint cascades, where local rules are applied repeatedly to derive held-out transitions in order of proof depth \(d\). Unlike general constraint propagation, a cascade predicts that transitions are inferred only after all their derivation premises, the other held-out transitions required for the derivation, are inferred (Figure \ref{fig:commutative_world}b). We examine \(150\) checkpoints for each of \(5\) seeds trained over composition lengths \(T=2,3,4\) on the commutative world to find the percentage of inferred transitions missing their derivation premises. To remove timing confounds (on average \(d=n+1\) is learned after \(d=n\)), we compare to a null which shuffles the identity of \(d=n\) transitions, preserving the timing but removing any dependency structure. We find that only \(3\%\) of correct inferences made by the transformer were made without having correctly inferred their derivation premises, compared to \(11\%\) for the null model, a difference of \(3.7\times\). Throughout training, the model's correct held-out predictions are almost always derivable from seen transitions and other predictions, satisfying the constraint cascade condition.

\subsection{Longer composition length enables deeper propagation}
Training on longer composition lengths \(T\) with the same number of training steps improves generalization at higher proof depths (Figure \ref{fig:commutative_world}c). Increasing \(T\) from \(2\) to \(3\) maintains \(d=1\) recovery at \(\geq 95\%\) but significantly improves \(d=2\) by \(+43\%\) and similarly improves recovery of \(3\leq d\leq 8\) with an average increase of \(+19\%\). Raising composition length \(T\) from \(3\) to \(4\) maintains \(d=1\) and \(d=2\) recovery while again accumulating improvement for \(3\leq d \leq 8\) over \(T=3\) by \(+14\%\). Decay curves also change qualitatively, from the geometric-like decay of \(T=2\), which is indistinguishable from random guessing by \(d=5\), to plateau-like behavior at \(T=3\) and \(T=4\), where generalization appears to flatten out at roughly \(10\%\) and \(20\%\), respectively. These results suggest training on longer composition lengths may be key to deep propagation of constraints.

\section{Discussion}
Identical data, learned under a different objective, can turn a model that fails to generalize into one that learns and propagates global constraints. Next-state and compositional objectives differ only in whether intermediate states are shown. Hiding them incentivizes a shared state representation (Section~\ref{sec:shared_space}), where a constraint seen in training becomes a relation between actions at every state and so propagates to held-out transitions. The next-state loss never favors the constraint, so it is neither learned nor propagated. Global understanding depends on how training combines local observations.

Training objective, perhaps more than architecture or domain, determines which constraints a model propagates. Under each objective, the same constraints are propagated across attention, recurrent, and state-space architectures. Constraints propagated compositionally in monoid worlds are also propagated in visual world models and LLMs, including at scale in scanned homes and naturalistic articles. This pattern suggests that controlled monoid worlds capture general learning biases that extend to richer tasks and may provide simple environments for designing better-propagating algorithms.

Propagation decides whether a world model supports control beyond the transitions it has seen. With search or imagination, compositional models reach goals through held-out transitions, while next-state mispredictions can make planning worse than using seen transitions alone. Policies trained on frozen compositional representations reach these goals, without search or training on held-out transitions, so propagated constraints are encoded in representation geometry. Changing only the objective also improves world-model benchmarks in physics, planning, and weather (Appendix \ref{appx:literature_benchmarks}).

Even when models propagate constraints, propagation can be shallow, with generalization decreasing sharply with proof depth. Higher proof depth transitions are also recovered later and more slowly. Together with training dynamics, this suggests an internal process resembling a slow, inaccurate constraint cascade. Increasing composition length improves generalization to deeper consequences, suggesting that propagation reach depends on the organization of local information it learns from. Propagating constraint consequences deeply is essential for consistent information integration. 

\section{Related work}
LLMs often learn facts without propagating their consequences. The reversal curse shows that training on one direction of a relation can fail to recover its reverse \citep{berglund2024reversal}. Models also compose relations poorly in implicit reasoning \citep{balesni2025lessonsstudyingtwohoplatent}. Training changes address individual failures such as reversal \citep{ma2026breakingreversalcurseautoregressive} and composition \citep{pmlr-v267-feng25m}, but remain specialized. We frame these failures and improvements as learning and propagating constraints.

Accurate local predictions do not imply coherent global understanding \citep{vafa2024worldmodel}. Prior work improves consistency with reverse-prediction training \citep{kwon2026stable} or modified architectures that recover group structure in synthetic environments \citep{pmlr-v202-keurti23a}. Our constraint propagation framework extends beyond groups to semantic constraints such as parent--child converse.

Compositional training was introduced for knowledge-graph embeddings, where it improves multi-hop queries and single-edge completion \citep{guu-etal-2015-traversing}. In planning, hiding previous steps removes a local shortcut in next-token training \citep{bachmann2025pitfallsnexttokenprediction}. We confirm that hiding intermediate states improves held-out accuracy in deterministic monoid worlds. Each held-out transition follows from the observed transitions and the world’s constraint. This lets us test which constraints each objective propagates and how recovery drops with proof depth. We analyze inverse propagation in a linear model and test whether models can use recovered transitions for control.

\section{Conclusion}
We frame global understanding as learning constraints from local observations and propagating their consequences. In monoid worlds, where seen transitions and constraints entail held-out transitions, next-state training fails to propagate non-trivial constraints across architectures, while compositional training improves propagation through a shared state representation. Improvements hold at scale in visual world models and LLMs, and enable control through held-out transitions. Propagation remains shallow, though longer composition lengths help. These results suggest that global understanding requires training that integrates local observations and propagates their consequences deeply.

\newpage

\subsection*{AI use statement}
In this work, we used generative AI tools to implement methods, including code that generates the vision and language worlds, extracts the Wikidata graphs, trains and evaluates the models, and to organize our code base. We also used an LLM to generate paragraphs for the synthetic training articles (Appendix \ref{appx:composition_scale}). Additionally, we used generative AI tools to improve readability, reorganize, suggest improvements, and proofread drafts. We have reviewed all AI-assisted work. We reviewed and tested all AI-written code and reviewed every AI-suggested edit to the text. We take responsibility for the final content of this work, including text, claims or artifacts produced with the aid of generative AI.

\subsection*{Reproducibility statement}
The supplementary material contains source code for replication, the AI2-THOR rooms with their rendered observations, and the extracted Wikidata graphs. Scripts in it regenerate every run configuration, including all architectures, hyperparameters, and seeds, and reproduce every figure and table.

\bibliography{references}
\bibliographystyle{iclr2027_conference}

\newpage
\appendix
\section{Scaling model, world, and compute under next-state training does not enable constraint propagation}
\label{appx:scale_next_state}
We explore whether scaling a transformer under next-state training can enable constraint propagation on the inverse world by individually scaling model width from \(64\) to \(320\) (\(0.33\)M to \(5.6\)M parameters), world size from \(1024\) to \(8192\) states, and training time from \(15\)k to \(300\)k steps. We find that every next-state trained model fits the observed transitions but fails to recover held-out transitions (all at random floor). The compositionally trained control does generalize, achieving \(97\%\) on held-out inverse transitions.

\begin{table}[h]
\caption{\textbf{The next-state objective does not propagate constraints even after scaling model, world, and compute.} Transformers trained on inverse world at \(T=2\). Every row is next-state training except the last, which trains the same world, training paths, and model but compositionally. Accuracies are average \(\pm\) standard deviation over five seeds.}
\label{tab:scale_next_state}
\begin{center}
\small
\begin{tabular}{lrrrrrr}
\toprule
 & Width & Parameters (M) & States ($N$) & Steps & Seen (\%) & Held-out (\%) \\
\midrule
Model size & 64 & 0.33 & 1024 & 150k & 100.0 & 0.4 $\pm$ 0.5 \\
 & 128 & 1.05 & 1024 & 150k & 100.0 & 0.4 $\pm$ 0.5 \\
 & 256 & 3.67 & 1024 & 150k & 100.0 & 0.2 $\pm$ 0.4 \\
 & 320 & 5.57 & 1024 & 150k & 100.0 & 0.4 $\pm$ 0.5 \\
\midrule
World size & 320 & 5.57 & 1024 & 150k & 100.0 & 0.4 $\pm$ 0.5 \\
 & 320 & 7.54 & 4096 & 150k & 100.0 & 0.1 $\pm$ 0.1 \\
 & 320 & 10.16 & 8192 & 150k & 100.0 & 0.1 $\pm$ 0.1 \\
\midrule
Training time & 320 & 10.16 & 8192 & 15k & 100.0 & 0.1 $\pm$ 0.1 \\
 & 320 & 10.16 & 8192 & 75k & 100.0 & 0.1 $\pm$ 0.1 \\
 & 320 & 10.16 & 8192 & 150k & 100.0 & 0.1 $\pm$ 0.1 \\
 & 320 & 10.16 & 8192 & 300k & 100.0 & 0.1 $\pm$ 0.1 \\
\midrule
Compositional control & 320 & 10.16 & 8192 & 150k & 100.0 & 97.0 $\pm$ 0.5 \\
\bottomrule
\end{tabular}
\end{center}
\end{table}

\section{Models incorrectly propagate copy constraints in conflict with seen transitions}
\label{appx:copy_bias_next_state}

\begin{figure}[h]
\centering
\includegraphics[width=\textwidth]{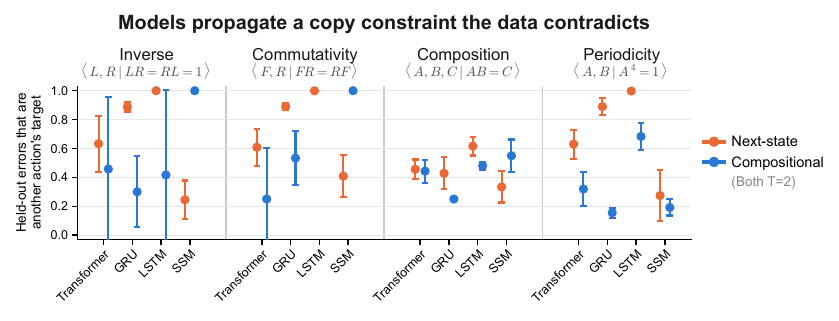}
\caption{\textbf{Models systematically fail by recovering held-out transitions with another action's target.} When models fail to recover a held-out transition \((s, A, A(s))\), they instead often predict \(B(s)\) for another action \(B\). Points averaged over 5 seeds and bars show standard deviations.}
\label{fig:copy_bias_next_state}
\end{figure}

When models trained with either the next-state or the compositional objective fail to propagate a constraint, they fail in a systematic way. For an incorrectly answered held-out transition \((s, A, A(s))\), we count a copy when the model predicts \(B(s)\), the target of another action \(B\) from the same state. Across inverse, commutative, composition, and periodicity worlds and all four architectures, \(64\%\) of next-state errors are copies, compared with \(36\%\) of compositional errors. Copy rate depends on architecture, with the highest rates (all next-state trained) being LSTMs at \(90\%\), GRUs at \(77\%\), and transformers at \(58\%\).

\section{Compositionally trained models propagate constraints in proportion to observed evidence}
\label{appx:evidence_scaling_compositional}
We vary constraint evidence, the number of seen transitions satisfying the constraint, while fixing the world size. We train a transformer at \(T=2\) over the composition world \(N=1024\), varying the amount of held-out \(C\)-transitions from \(10\%\) to \(97\%\) (evidence levels of 922 and 31, respectively). Next-state trained models do not generalize, while compositionally trained models generalize monotonically with evidence: \(2\%\), \(4\%\), \(11\%\), \(24\%\), \(62\%\), and \(93\%\) at evidence 31, 62, 128, 256, 512, and 922.

\begin{figure}[h]
\centering
\includegraphics[width=0.8\textwidth]{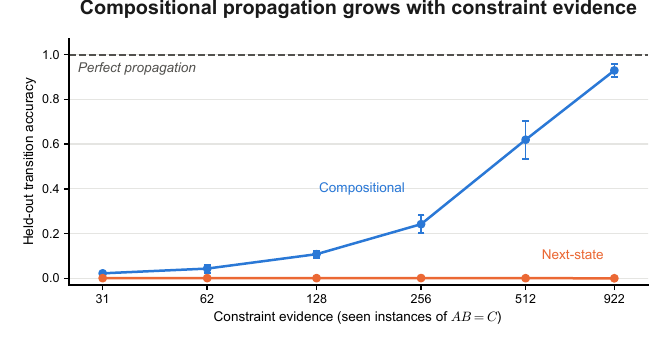}
\caption{\textbf{Compositionally trained transformers recover held-out transitions in proportion to constraint evidence.} Transformers trained on the composition world \(N=1024\) with \(T=2\) paths. Points averaged over 5 seeds and bars show standard deviations.}
\label{fig:evidence_scaling_compositional}
\end{figure}

\section{Composition length has minimal effect on recovery of immediately derivable (proof depth \(d=1\)) transitions}
\label{appx:composition_length}
\begin{figure}[h]
\centering
\includegraphics[width=\textwidth]{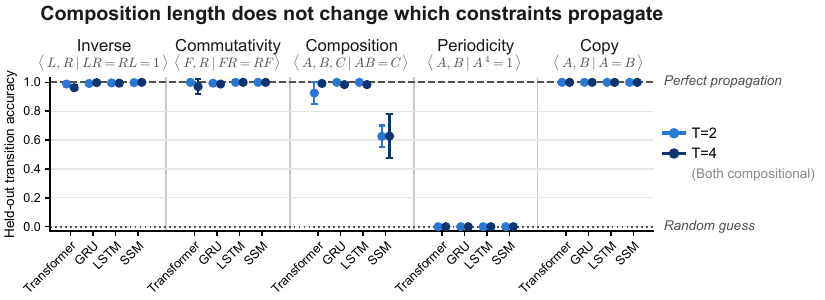}
\caption{\textbf{Composition length has minimal effect on propagating immediately derivable transitions.} Compositional training at \(T=4\) propagates inverse, commutative, and composition constraints with an average accuracy of \(96\%\), the same as at \(T=2\), and copy at \(100\%\). Increasing composition length does not rescue periodicity. Points averaged over 5 seeds and bars show standard deviations.}
\label{fig:composition_length}
\end{figure}

We explore whether increasing composition length improves constraint propagation to held-out transitions that are immediately derivable from seen ones (\(d=1\)). We train every compositional model of Figure \ref{fig:arch_relations} at \(T=4\) in addition to \(T=2\), keeping the worlds, held-out transitions, and \(150\)k training steps fixed. Compositional path length is the only difference. All held-out transitions have proof depth \(d=1\). Compositionally trained models at \(T=4\) propagate inverse, commutative, and composition constraints with an average accuracy of \(96\%\), the same as at \(T=2\), and copy at \(100\%\) (Figure \ref{fig:composition_length}). Neither \(T=2\) nor \(T=4\) propagates periodicity.

\section{Branching acts as noise on the inverse constraint}
\label{appx:branching}
\begin{figure}[h]
\centering
\includegraphics[width=0.8\textwidth]{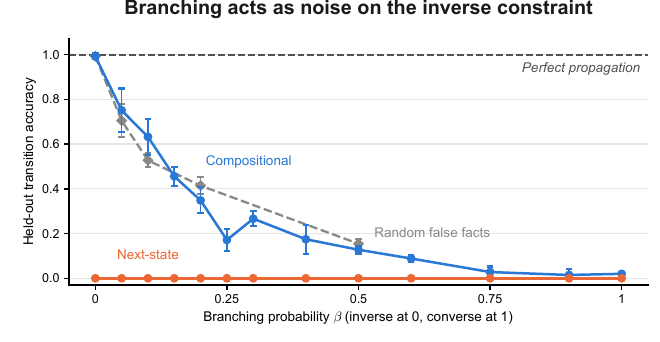}
\caption{\textbf{Branching acts as noise on the inverse constraint.} Compositional training propagates the inverse constraint (\(\beta=0\)) at \(99\%\), but only a \(10\%\) chance of a second child (\(\beta=0.1\)) drops generalization to \(63\%\), and for \(\beta\geq0.75\) accuracy drops to \(\leq 3\%\). Random false child facts that violate the inverse constraint as much as branching (grey) give the same accuracy. Next-state training fails at every \(\beta\). Points averaged over 3 seeds and bars show standard deviations.}
\label{fig:branching}
\end{figure}

We explore how branching affects converse constraint propagation by building synthetic family trees that interpolate between purely inverse \(\beta=0\) and purely converse constraints \(\beta=1\). The tree is generated breadth-first with \(N=1024\), where each parent has one guaranteed child plus another with probability \(\beta\) (expected number of children is \(1+\beta\)). We hold out \(52\) parent and \(52\) child relationships (\(5\%\)) which are jointly entailed by observed transitions and the converse constraint. We train a transformer on \(T=2\) paths for 150k steps with the compositional or the next-state objective on branching worlds of various \(\beta\). Compositional training on the line world (pure inverse \(\beta=0\)) generalizes at \(99\%\) accuracy, but generalization drops to \(63\%\) at \(\beta=0.1\), \(35\%\) at \(\beta=0.2\), \(13\%\) at \(\beta=0.5\), and \(\leq 3\%\) for \(\beta \geq 0.75\) (Figure \ref{fig:branching}). Next-state training fails for all \(\beta\). 

Branching acts as noise on the inverse constraint. In probabilistic worlds, noise that preserves the constraints leaves propagation intact (Appendix \ref{appx:probabilistic}), but with branching the parent's child is the starting entity only part of the time and otherwise its sibling. To test whether inverse violation explains the drop, we add random false child facts \((x, \text{child}, w)\), with \(w\) a random entity, to the line world. At each \(\beta\), we choose the number of false facts so that the L2 deviation of \(P_{\text{parent}}P_{\text{child}}\) and \(P_{\text{child}}P_{\text{parent}}\) from \(I\) is the same as in the branching world. Accuracy matches branching: \(71\%\), \(53\%\), \(41\%\) and \(15\%\) against \(75\%\), \(63\%\), \(35\%\) and \(13\%\) at \(\beta = 0.05, 0.1, 0.2, 0.5\) (Figure \ref{fig:branching}).

\section{Composition propagation occurs without path shortcuts}
\label{appx:composition_shortcut}
We explore whether composition \(AB=C\) propagates when shortcut paths \(s \xrightarrow{A} A(s) \xrightarrow{B} C(s)\) for a held-out transition \((s, C, C(s))\) are removed from training. We consider the composition monoid world \(N=1024\) and hold out \(10\%\) of \(C\)-transitions. We train a transformer on \(T=2\) paths for 150k steps under the next-state or compositional objective with and without shortcut paths in the training dataset. Next-state trained models fail in both settings. Compositionally trained models generalize at \(92\%\) accuracy with shortcut paths and \(50\%\) without (Table \ref{tab:composition_shortcut}). 

\begin{table}[h]
\caption{\textbf{Compositional training propagates composition without shortcut paths.} Transformers trained on the composition world \(N=1024\) at \(T=2\) with and without the shortcut path \(s \xrightarrow{A} A(s) \xrightarrow{B} C(s)\) of each held-out transition. Accuracies are average \(\pm\) standard deviation over five seeds.}
\label{tab:composition_shortcut}
\begin{center}
\small
\begin{tabular}{llrr}
\toprule
Objective & Shortcut paths & Seen (\%) & Held-out (\%) \\
\midrule
Next-state & with & 100 & 0 $\pm$ 0 \\
 & without & 100 & 0 $\pm$ 0 \\
\midrule
Compositional & with & 100 & 92 $\pm$ 6 \\
 & without & 100 & 50 $\pm$ 10 \\
\bottomrule
\end{tabular}
\end{center}
\end{table}

\section{Periodicity is not propagated across composition lengths}
\label{appx:periodicity}
Composition length is varied \(1\leq T\leq 8\) and \(A^k=1\) periodicity length \(k=3, 4, 5\) over \(N=1020\) periodicity worlds to see if compositional training can propagate the constraint. A training path contains a full wrap-around loop when \(T \geq k\). All held-out transitions have proof depth \(d=1\). We train transformers with the compositional objective for 150k steps. Every model fits observed transitions but fails to propagate the periodic constraint (all at random floor) even with \(T \geq k\) (Table \ref{tab:periodicity}).

\begin{table}[h]
\caption{\textbf{Periodicity is not propagated across composition lengths.} All transformers trained compositionally on \(N=1020\) periodicity worlds with \(A^k=1\) for \(k=3,4,5\) and composition lengths \(1\leq T\leq 8\) fail to propagate the periodicity constraint. Accuracies are averaged over five seeds.}
\label{tab:periodicity}
\begin{center}
\small
\setlength{\tabcolsep}{4pt}
\begin{tabular}{lrrrrrrrr}
\toprule
 & \multicolumn{8}{c}{Held-out (\%) at path length $T$} \\
\cmidrule(l){2-9}
Constraint & 1 & 2 & 3 & 4 & 5 & 6 & 7 & 8 \\
\midrule
$A^3=1$ & 0.4 $\pm$ 0.5 & 0.4 $\pm$ 0.5 & 0.2 $\pm$ 0.4 & 0.0 $\pm$ 0.0 & 0.0 $\pm$ 0.0 & 0.0 $\pm$ 0.0 & 0.0 $\pm$ 0.0 & 0.0 $\pm$ 0.0 \\
$A^4=1$ & 0.6 $\pm$ 0.5 & 0.2 $\pm$ 0.4 & 0.2 $\pm$ 0.4 & 0.2 $\pm$ 0.4 & 0.0 $\pm$ 0.0 & 0.2 $\pm$ 0.4 & 0.0 $\pm$ 0.0 & 0.2 $\pm$ 0.4 \\
$A^5=1$ & 0.6 $\pm$ 0.5 & 0.2 $\pm$ 0.4 & 0.0 $\pm$ 0.0 & 0.0 $\pm$ 0.0 & 0.0 $\pm$ 0.0 & 0.0 $\pm$ 0.0 & 0.0 $\pm$ 0.0 & 0.0 $\pm$ 0.0 \\
\bottomrule
\end{tabular}
\end{center}
\end{table}

We also train for 1.5M steps, ten times longer, to test for grokking-like delays. On the periodicity world of Section~\ref{sec:monoid_world}, transformers at \(T=2\) and \(T=4\) and GRU, LSTM, and SSM models at \(T=4\) fit every seen transition within 6k steps but still fail to propagate the periodic constraint (Table~\ref{tab:periodicity_long}).

\begin{table}[h]
\caption{\textbf{Training ten times longer does not propagate periodicity.} Held-out accuracy (\%) of models trained compositionally on the periodicity world of Section~\ref{sec:monoid_world}. Accuracies are averaged over three seeds.}
\label{tab:periodicity_long}
\begin{center}
\small
\begin{tabular}{llrr}
\toprule
Model & \(T\) & 150k steps & 1.5M steps \\
\midrule
Transformer & 2 & 0.0 $\pm$ 0.0 & 0.0 $\pm$ 0.0 \\
Transformer & 4 & 0.0 $\pm$ 0.0 & 0.0 $\pm$ 0.0 \\
GRU & 4 & 0.0 $\pm$ 0.0 & 0.3 $\pm$ 0.6 \\
LSTM & 4 & 0.0 $\pm$ 0.0 & 0.3 $\pm$ 0.6 \\
SSM & 4 & 0.0 $\pm$ 0.0 & 0.0 $\pm$ 0.0 \\
\bottomrule
\end{tabular}
\end{center}
\end{table}

\section{Constraint propagation in probabilistic worlds}
\label{appx:probabilistic}
Probabilistic worlds replace action functions with Markov kernels. The states are \(\mathcal S:=[N]\) as before, but each action \(a\in\mathcal A\) is an \(N\times N\) stochastic matrix \(P_a\), where \(P_a(s,s')\) is the probability that \(a\) takes state \(s\) to \(s'\). Action sequences compose as Markov kernels, with matrix products, \(P_{ab}:=P_aP_b\), and with the identity matrix form a monoid. A monoid world is the special deterministic case where each row \(P_a(s,\cdot)\) puts all its mass on \(a(s)\). Local observations are sampled transitions \((s,a,s')\) with \(s'\sim P_a(s,\cdot)\).

The global constraints are equalities of kernel compositions: \(u=v\) when \(P_u=P_v\). In other words, at every initial state, the two action sequences reach the same distribution over terminal states. Certain constraints, like inverse and periodicity, force deterministic kernels. We study commutativity and composition. Semantic entailment carries over with a held-out row \(P_a(s,\cdot)\) entailed if all Markov kernels that satisfy the constraint and agree on the seen rows also agree on it.

We explore two \(\epsilon\)-parameterized action kernels \(P_a^\epsilon\) derived by adding noise to deterministic action maps \(M_a\), written as matrices with \(M_a(s,a(s))=1\). The first, uniform noise, follows the action map with probability \(1-\epsilon\) and otherwise moves to a uniformly random state: \(P_a^\epsilon=(1-\epsilon)M_a+\epsilon U\), where every entry of \(U\) is \(1/N\). The second, absorbing noise, follows the action map with probability \(1-\epsilon\) and otherwise moves to a new state \(s_\ast\) that every action leaves in place: \(P_a^\epsilon=(1-\epsilon)M_a+\epsilon F\), where every row of \(F\) puts all its mass on \(s_\ast\) and \(M_a(s_\ast,s_\ast)=1\), so the states become \(\mathcal S\cup\{s_\ast\}\). Commutativity \(P_A^\epsilon P_B^\epsilon=P_B^\epsilon P_A^\epsilon\) holds exactly at every \(\epsilon\). For composition, we define \(C\) as the composite \(P_C^\epsilon:=P_A^\epsilon P_B^\epsilon\). At \(\epsilon=0\), both reduce to the monoid worlds of Section \ref{sec:monoid_world}.

We keep the \(N=1024\) worlds and held-out transitions of Section \ref{sec:monoid_world}, which remain entailed. Training paths sample transitions \(s_{t+1}\sim P_{a_t}^\epsilon(s_t,\cdot)\), and no training path takes a held-out transition. We train transformers of width \(256\) on \(T=4\) paths. Gradients are noisy, so the learning rate is constant for 150k steps and then decays linearly to zero over 20k steps. For a held-out transition \((s,a)\), we report accuracy, whether \(a(s)\) is the most probable predicted state in \(\mathcal S\), and probability recovered, the predicted probability of \(a(s)\) divided by its true probability \(P_a^\epsilon(s,a(s))\). At \(\epsilon=0\), accuracy is the same as in Section \ref{sec:monoid_world}.

Next-state training fails to propagate commutativity and composition under both noises (Tables \ref{tab:absorbing} and \ref{tab:uniform}). Compositional training achieves \(99\)--\(100\%\) accuracy at every \(\epsilon\). It recovers most of the held-out transitions' probability at low noise, \(87\)--\(99\%\) at \(\epsilon=1/10\) and \(77\)--\(88\%\) at \(\epsilon=3/10\), and does moderately well at high noise, \(25\)--\(70\%\) at \(\epsilon=1/2\).

\begin{table}[h]
\caption{\textbf{Compositional training propagates constraints under absorbing noise, while next-state training does not.} Transformers trained on \(T=4\) paths in worlds where each action moves to the absorbing state \(s_\ast\) with probability \(\epsilon\). Accuracy is the percentage of held-out transitions \((s,a)\) whose most probable predicted state in \(\mathcal S\) is \(a(s)\). Probability is the predicted probability of \(a(s)\) as a percentage of its true probability. One seed each.}
\label{tab:absorbing}
\begin{center}
\small
\begin{tabular}{lrrrrr}
\toprule
 & & \multicolumn{2}{c}{Next-state} & \multicolumn{2}{c}{Compositional} \\
\cmidrule(lr){3-4}\cmidrule(l){5-6}
Constraint & \(\epsilon\) & Accuracy (\%) & Probability (\%) & Accuracy (\%) & Probability (\%) \\
\midrule
Commutativity & 0 & 1 & 1 & 100 & 100 \\
 & 1/10 & 1 & 0 & 100 & 97 \\
 & 3/10 & 1 & 0 & 100 & 77 \\
 & 1/2 & 1 & 0 & 100 & 25 \\
\midrule
Composition & 0 & 1 & 1 & 100 & 100 \\
 & 1/10 & 1 & 0 & 100 & 87 \\
 & 3/10 & 0 & 0 & 100 & 77 \\
 & 1/2 & 0 & 0 & 100 & 70 \\
\bottomrule
\end{tabular}
\end{center}
\end{table}

\begin{table}[h]
\caption{\textbf{Compositional training propagates constraints under uniform noise, while next-state training does not.} Transformers trained on \(T=4\) paths in worlds where each action moves to a uniformly random state with probability \(\epsilon\). Accuracy and probability are as in Table \ref{tab:absorbing}. One seed each.}
\label{tab:uniform}
\begin{center}
\small
\begin{tabular}{lrrrrr}
\toprule
 & & \multicolumn{2}{c}{Next-state} & \multicolumn{2}{c}{Compositional} \\
\cmidrule(lr){3-4}\cmidrule(l){5-6}
Constraint & \(\epsilon\) & Accuracy (\%) & Probability (\%) & Accuracy (\%) & Probability (\%) \\
\midrule
Commutativity & 0 & 0 & 0 & 100 & 100 \\
 & 1/10 & 1 & 0 & 100 & 93 \\
 & 3/10 & 1 & 0 & 100 & 84 \\
 & 1/2 & 1 & 1 & 99 & 55 \\
\midrule
Composition & 0 & 0 & 0 & 100 & 100 \\
 & 1/10 & 0 & 0 & 100 & 99 \\
 & 3/10 & 0 & 0 & 100 & 88 \\
 & 1/2 & 0 & 0 & 100 & 69 \\
\bottomrule
\end{tabular}
\end{center}
\end{table}

\section{Composition propagation at scale in naturalistic articles}
\label{appx:composition_scale}
We test whether composition constraint propagation by compositional training occurs when the Wikidata world is larger and the training text reads like naturalistic articles instead of templated sentences. We extend the family world of Section \ref{subsec:llm} to \(1{,}019\) and \(4{,}086\) renamed people and hold out \(8\%\) of grandparent facts. Each person is described by a Wikipedia-style article containing a career paragraph, written by an LLM from the Wikidata facts on another real person, and family sentences. Compositional articles use compositional phrasing for family relations while next-state articles use next-state phrasing. We evaluate with question and answer interactions. To teach the question format to base models, both objectives include training question examples about parents and seen grandparents of a random half of the people, and held-out grandparents never appear in them. Examples of articles, questions, and prompts are shown in Table \ref{tab:kinship_example}. We finetune Qwen3 base models (4B, 8B, and 14B) for 4k steps on \(1{,}019\) people and 16k steps on \(4{,}086\) people. Both objectives answer seen questions at \(84\)--\(99\%\). Next-state trained models fail on held-out questions (\(\leq 0.6\%\)), while compositionally trained models generalize at \(87\)--\(90\%\) on \(1{,}019\) people and \(73\)--\(75\%\) on \(4{,}086\) people (Table \ref{tab:kinship_scale}).

\begin{table}[h]
\caption{\textbf{Example training documents and test prompt.} Charity's grandparents are held out and Jeremy's are seen. Articles of the two objectives differ only in the phrasing of family sentences.}
\label{tab:kinship_example}
\begin{center}
\small
\begin{tabular}{p{0.2\textwidth}p{0.74\textwidth}}
\toprule
Compositional article & Charity Hopkins (1817--1898) was a French nun best known for her religious service and dedication to charitable work. She was born in Metz and died in Auteuil. \dots{} Charity Hopkins was the daughter of Sophie Lane and Harry Choi. Charity Hopkins's mother's father was Andrew Ling. \dots \\
\midrule
Next-state article & \dots{} Charity Hopkins was the daughter of Sophie Lane and Harry Choi. Charity Hopkins's mother was Sophie Lane, whose father was Andrew Ling. \dots \\
\midrule
Question document & Q: Who was Jeremy Cross's maternal grandfather? \newline A: Kevin Grey. \\
\midrule
Test prompt & Q: Who was Charity Hopkins's maternal grandfather? \newline A: \\
\bottomrule
\end{tabular}
\end{center}
\end{table}

\begin{table}[h]
\caption{\textbf{Compositional training propagates composition in naturalistic articles at larger scale.} Qwen3 models finetuned on Wikipedia-style articles and asked about seen and held-out grandparents. One seed each.}
\label{tab:kinship_scale}
\begin{center}
\small
\begin{tabular}{llrrrr}
\toprule
 & & \multicolumn{2}{c}{Next-state} & \multicolumn{2}{c}{Compositional} \\
\cmidrule(lr){3-4}\cmidrule(l){5-6}
People & Model & Seen (\%) & Held-out (\%) & Seen (\%) & Held-out (\%) \\
\midrule
1{,}019 & 4B & 95.3 & 0.0 & 94.3 & 86.7 \\
 & 8B & 97.7 & 0.6 & 99.0 & 89.1 \\
 & 14B & 93.7 & 0.0 & 94.3 & 89.7 \\
\midrule
4{,}086 & 4B & 89.0 & 0.0 & 89.7 & 75.2 \\
 & 8B & 89.7 & 0.0 & 93.0 & 73.4 \\
 & 14B & 84.0 & 0.1 & 88.0 & 73.8 \\
\bottomrule
\end{tabular}
\end{center}
\end{table}

\section{Geometric constraint propagation at scale in real 3D scans}
\label{appx:geometric_scale}

\begin{table}[h]
\caption{\textbf{Setup of visual world model studies.} AI2-THOR (Section \ref{sec:vision_world}) compared with Habitat.}
\label{tab:habitat_scale}
\begin{center}
\small
\begin{tabular}{lll}
\toprule
 & AI2-THOR (Section \ref{sec:vision_world}) & Habitat \\
\midrule
Environment & generated ProcTHOR rooms & scanned real homes (HM3D) \\
States & \(1{,}758\) in \(2\) rooms & \(30{,}496\) in \(12\) houses \\
Observation & \(32\times32\) RGB & \(64\times64\) RGB \\
Actions & \(4\) steps at one heading & \(4\) steps relative to heading, \(2\) turns \\
Constraints & inverse, commutative & inverse, commutative, periodicity, rotation \\
Path length & \(T=2\) & \(T=4\) \\
World model & \(2.9\)M parameters (width \(64\), \(4\) layers) & \(174\)M parameters (width \(1024\), \(12\) layers) \\
\bottomrule
\end{tabular}
\end{center}
\end{table}

We test whether geometric constraint propagation by compositional training holds at scale in environment, model, and actions (Table \ref{tab:habitat_scale}).

Houses are HM3D scans of real homes \citep{ramakrishnan2021hm3d} rendered in Habitat \citep{savva2019habitat}. A state is the floor position with one of four headings, observed as a \(64\times64\) RGB view. The agent steps (F)orward/(B)ackward/(L)eft/(R)ight relative to its heading and turns \(90^\circ\) (\(T_L\)/\(T_R\)). A collision leaves the state unchanged. On open floor the moves satisfy inverse (\(FB=LR=T_LT_R=1\)), commutative (\(FR=RF\)), periodicity (\(T_R^4=1\)), and rotation (\(FT_R=T_RL\)) constraints.

A \(174\)M-parameter transformer (width \(1024\), \(12\) layers) is trained for 150k steps on all length \(T=4\) paths of seen transitions in 12 houses, predicting next observations or states. Six houses show every transition. Each of five test houses shows only the actions of one constraint and holds out up to \(128\) transitions entailed by that constraint alone, choosing those whose next observation is least visible from the current one to minimize visual confounds. A sixth house is explored by a random policy and holds out the transitions the constraints entail from its explored ones.

Compositional training propagates the constraints at an average of \(66\%\) from observations and \(69\%\) from states, against \(13\%\) and \(6\%\) for next-state training. Unlike in the monoid periodicity worlds (Appendix \ref{appx:periodicity}), periodicity is propagated due to a commuting second action (Appendix \ref{appx:linear_periodicity}).

\begin{table}[h]
\caption{\textbf{Compositional training propagates geometric constraints at scale in real 3D scans.} Accuracy (\%) on held-out transitions of transformers trained on 12 HM3D houses, mean \(\pm\) standard deviation over three seeds.}
\label{tab:habitat}
\begin{center}
\small
\setlength{\tabcolsep}{3pt}
\begin{tabular}{llrrrrrr}
\toprule
 & & & & \multicolumn{2}{c}{Observation} & \multicolumn{2}{c}{State} \\
\cmidrule(lr){5-6}\cmidrule(l){7-8}
House & Constraint & Held & Baseline & Next-state & Compositional & Next-state & Compositional \\
\midrule
Inverse (steps) & \(LR=1\) & 128 & 15.6 & 10 \(\pm\) 2 & 51 \(\pm\) 4 & 0 \(\pm\) 0 & 40 \(\pm\) 10 \\
Inverse (turns) & \(T_LT_R=1\) & 128 & - & 1.8 \(\pm\) 0.5 & 71 \(\pm\) 4 & 0 \(\pm\) 0 & 87 \(\pm\) 2 \\
Commutative & \(FR=RF\) & 69 & 20.3 & 28 \(\pm\) 3 & 62 \(\pm\) 5 & 0 \(\pm\) 0 & 22 \(\pm\) 2 \\
Periodicity & \(T_R^4=1\) & 128 & - & 0.3 \(\pm\) 0.5 & 46 \(\pm\) 7 & 30 \(\pm\) 10 & 100 \(\pm\) 0 \\
Rotation & \(FT_R=T_RL\) & 128 & - & 26 \(\pm\) 4 & 98 \(\pm\) 2 & 0 \(\pm\) 0 & 98 \(\pm\) 2 \\
\midrule
Explored & all & 2.6k & -- & 29.2 \(\pm\) 0.7 & 88.7 \(\pm\) 0.3 & 10 \(\pm\) 2 & 85.6 \(\pm\) 0.2 \\
\bottomrule
\end{tabular}
\end{center}
\end{table}

For Figure \ref{fig:control}, a separate model learns from one explored house alone with \(588\) random-policy episodes of \(20\) moves covering \(57\%\) of transitions.

\section{Linear models reproduce the constraint propagation of nonlinear models}
\label{appx:linear_evidence}
\label{appx:linear_theory}
We explore whether a linear model captures the constraint learning and propagation of our nonlinear models. The linear model represents each state \(s\) by a learned vector \(x_s\) and each action by a matrix \(M_a\), and reads out the next state from \(U M_a x_s\). We train it with width 32 on the worlds of Figure \ref{fig:arch_relations} at \(T=2\) under both objectives. The linear model reproduces the nonlinear results: compositional training propagates inverse, commutative, and composition constraints, next-state training propagates only copy, and neither propagates periodicity (Table \ref{tab:linear_model_accuracy}). Linear probes of the nonlinear models, one matrix per action fit on seen transitions with embeddings and output layer frozen, match their held-out accuracy on inverse worlds: \(98.7\%\) against \(99.0\%\) for compositional models and \(0.3\%\) against \(0.4\%\) for next-state models.

\begin{table}[h]
\caption{\textbf{A trained linear model reproduces the constraint propagation of nonlinear models.} Held-out accuracy (\%) at \(N=1024\) and \(T=2\). Nonlinear models are the four architectures of Figure \ref{fig:arch_relations}. Accuracies are average \(\pm\) standard deviation over five seeds.}
\label{tab:linear_model_accuracy}
\begin{center}
\small
\begin{tabular}{lrrrr}
\toprule
 & \multicolumn{2}{c}{Linear model} & \multicolumn{2}{c}{Nonlinear models} \\
\cmidrule(lr){2-3}\cmidrule(l){4-5}
World & Compositional & Next-state & Compositional & Next-state \\
\midrule
Inverse & 92.7 $\pm$ 0.9 & 0.0 $\pm$ 0.0 & 99.4 $\pm$ 0.9 & 0.5 $\pm$ 0.6 \\
Commutativity & 94.5 $\pm$ 0.9 & 0.8 $\pm$ 1.8 & 99.9 $\pm$ 0.5 & 0.7 $\pm$ 0.8 \\
Composition & 75.1 $\pm$ 0.5 & 0.0 $\pm$ 0.0 & 88.8 $\pm$ 16.6 & 0.1 $\pm$ 0.3 \\
Periodicity & 0.0 $\pm$ 0.0 & 0.0 $\pm$ 0.0 & 0.0 $\pm$ 0.0 & 0.1 $\pm$ 0.5 \\
Copy & 99.7 $\pm$ 0.3 & 100.0 $\pm$ 0.0 & 100.0 $\pm$ 0.0 & 100.0 $\pm$ 0.0 \\
\bottomrule
\end{tabular}
\end{center}
\end{table}

\section{Shared state representations propagate the inverse constraint}
\label{appx:linear_completion}
We show that a linear model with a shared state representation, where \(M_a x_u = x_{a(u)}\) on every seen transition \((u, a, a(u))\), recovers a held-out transition \((s, R, t)\) whose reverse \((t, L, s)\) is seen. Every state \(u\) with a seen path out by \(L\) and back by \(R\) satisfies \(M_R M_L x_u = x_u\). Since states far outnumber dimensions (\(N=1024 \gg d=32\)), the vectors of these states generically span the entire space, so \(M_R M_L = I\). The seen reverse gives \(x_s = M_L x_t\), so the model outputs
\begin{equation}
    M_R x_s = M_R M_L x_t = x_t ,
    \label{eq:inverse_completion}
\end{equation}
the same vector as on the seen transition into \(t\), recovering the held-out transition \((s, R, t)\).

Compositional training, unlike next-state training, incentivizes shared state representations. It fits \(U M_b M_a x_u\) and \(U M_b x_{a(u)}\) to the same target, and the loss falls when two predictions with the same target are replaced by their average. Next-state training fits only \(U M_a x_u\), which is unchanged when every \(M_a\) is replaced by \(G M_a\) and \(U\) by \(U G^{-1}\) for any invertible \(G\), although the outputs become \(G M_a x_u\). Its objective does not favor shared state representations. 

Trained models approximately share the representation under compositional training and not under next-state training (Appendix \ref{appx:shared_representation}). On periodicity worlds an unconstrained second action \(B\) weakens the shared representation and propagation fails, and both are restored when \(B\) is the identity or commutes with the periodic action (Appendix \ref{appx:linear_periodicity}).

\section{Compositional training learns a shared state representation}
\label{appx:shared_representation}
We test whether trained models satisfy the two equations of Appendix \ref{appx:linear_completion}: the shared state representation \(M_a x_u = x_{a(u)}\), and the constraint as an equation between action matrices, such as \(M_R M_L = I\). In the linear models of Table \ref{tab:linear_model_accuracy} we read both from the weights. For the shared representation we report the cosine between \(M_a x_u\) and \(x_{a(u)}\) over seen transitions and over held-out transitions. For the constraint we report the cosine between its two sides as matrices (scale invariant). 

Under compositional training both equations hold approximately in every world. Under next-state training neither holds, except the copy constraint \(M_A = M_B\), the only constraint it propagates (Table \ref{tab:shared_representation}). On inverse, commutativity, and copy worlds the shared representation extends to held-out transitions. On periodicity worlds it does not, and the constraint is not propagated (Appendix \ref{appx:linear_periodicity}). On composition worlds it does not either, but \(M_B M_A = M_C\) turns each held-out transition by \(C\) into a seen path by \(A\) then \(B\).

\begin{table}[h]
\caption{\textbf{Compositional training learns the shared state representation and the constraint between action matrices, and next-state training learns only the copy constraint.} Trained linear models at \(N=1024\) and \(T=2\). Seen and held-out are the cosine between \(M_a x_u\) and \(x_{a(u)}\) over seen and over held-out transitions. Under next-state training the held-out cosine is at most 0.02 in every world. Constraint is the cosine between the two sides of the constraint as matrices: \(M_R M_L = I\) and \(M_L M_R = I\), \(M_R M_F = M_F M_R\), \(M_B M_A = M_C\), \(M_A^4 = I\), and \(M_A = M_B\). Averages \(\pm\) standard deviation over five seeds.}
\label{tab:shared_representation}
\begin{center}
\small
\begin{tabular}{lrrrrr}
\toprule
 & \multicolumn{3}{c}{Compositional} & \multicolumn{2}{c}{Next-state} \\
\cmidrule(lr){2-4}\cmidrule(l){5-6}
World & Seen & Held-out & Constraint & Seen & Constraint \\
\midrule
Inverse & 0.89 $\pm$ 0.00 & 0.83 $\pm$ 0.00 & 0.87 $\pm$ 0.01 & 0.00 $\pm$ 0.01 & 0.00 $\pm$ 0.03 \\
Commutativity & 0.84 $\pm$ 0.01 & 0.78 $\pm$ 0.01 & 0.88 $\pm$ 0.01 & 0.00 $\pm$ 0.00 & 0.03 $\pm$ 0.03 \\
Composition & 0.68 $\pm$ 0.00 & 0.34 $\pm$ 0.01 & 0.94 $\pm$ 0.00 & $-$0.01 $\pm$ 0.01 & $-$0.02 $\pm$ 0.03 \\
Periodicity & 0.71 $\pm$ 0.00 & 0.22 $\pm$ 0.02 & 0.80 $\pm$ 0.03 & 0.00 $\pm$ 0.01 & 0.07 $\pm$ 0.05 \\
Copy & 0.82 $\pm$ 0.00 & 0.81 $\pm$ 0.00 & 0.93 $\pm$ 0.01 & 0.00 $\pm$ 0.00 & 0.92 $\pm$ 0.01 \\
\bottomrule
\end{tabular}
\end{center}
\end{table}

Nonlinear models have no action matrices in their weights, so we fit one matrix per action that maps each state's embedding to its successor's. We fit it on 80\% of the seen transitions and report the cosine between its output and the successor's embedding on the remaining seen transitions and on the held-out transitions. The pattern is the same as in the linear models: under compositional training the matrices fit the seen transitions and extend to the held-out transitions of inverse, commutativity, and copy worlds, and under next-state training they extend to none (Table \ref{tab:shared_representation_nonlinear}). A fitted matrix shows that such a matrix exists, not that the model uses it, so it can overestimate the shared representation. In the linear models it agrees with the model's own matrices to within 0.03, except on seen transitions under next-state training on composition worlds, where it gives 0.53 against \(-0.01\). The nonlinear value of 0.40 there is likely such an overestimate.

\begin{table}[h]
\caption{\textbf{Nonlinear models learn a shared state representation under compositional training and not under next-state training.} One matrix per action is fitted to the state embeddings of the nonlinear models of Table \ref{tab:linear_model_accuracy} on 80\% of the seen transitions, at \(N=1024\) and \(T=2\). Seen and held-out are the cosine between its output and the successor's embedding on the remaining seen transitions and on the held-out transitions. Averages \(\pm\) standard deviation over four architectures and five seeds.}
\label{tab:shared_representation_nonlinear}
\begin{center}
\small
\begin{tabular}{lrrrr}
\toprule
 & \multicolumn{2}{c}{Compositional} & \multicolumn{2}{c}{Next-state} \\
\cmidrule(lr){2-3}\cmidrule(l){4-5}
World & Seen & Held-out & Seen & Held-out \\
\midrule
Inverse & 0.97 $\pm$ 0.01 & 0.89 $\pm$ 0.03 & 0.12 $\pm$ 0.06 & 0.09 $\pm$ 0.05 \\
Commutativity & 0.92 $\pm$ 0.03 & 0.87 $\pm$ 0.03 & 0.08 $\pm$ 0.03 & 0.05 $\pm$ 0.03 \\
Composition & 0.64 $\pm$ 0.01 & 0.28 $\pm$ 0.03 & 0.40 $\pm$ 0.06 & 0.01 $\pm$ 0.01 \\
Periodicity & 0.67 $\pm$ 0.02 & 0.10 $\pm$ 0.02 & 0.01 $\pm$ 0.02 & 0.00 $\pm$ 0.01 \\
Copy & 0.88 $\pm$ 0.06 & 0.88 $\pm$ 0.06 & 0.05 $\pm$ 0.03 & 0.04 $\pm$ 0.02 \\
\bottomrule
\end{tabular}
\end{center}
\end{table}

\section{Periodicity propagation fails due to a poor shared state representation}
\label{appx:linear_periodicity}
A held-out periodicity transition \((s_3, A, s_0)\) closes a cycle \(s_0 \to s_1 \to s_2 \to s_3\) whose three transitions are seen. With a shared state representation, the intact cycles give \(M_A^4 = I\) and the seen transitions give \(x_{s_3} = M_A^3 x_{s_0}\), so the model outputs
\begin{equation}
    M_A x_{s_3} = M_A^4 x_{s_0} = x_{s_0} ,
    \label{eq:periodic_completion}
\end{equation}
recovering the held-out transition \((s_3, A, s_0)\). Compositional training, however, does not learn a suitable shared state representation on periodicity worlds.

Periodicity is the only constraint in our worlds that involves a single action. Its held-out transition \((s_3, A, s_0)\) is the only transition by \(A\) out of \(s_3\) and into \(s_0\), so without another action \(s_3\) is never an input and \(s_0\) is never a target. Transformers trained compositionally on \(\langle A \mid A^4 = 1\rangle\) recover none of the held-out transitions. A second action \(B\) trains both vectors (like in other constraints). In our periodicity world (Section \ref{sec:monoid_world}), \(B\) is an arbitrary permutation, and \(A\) and \(B\) together can map any two states to any other two. A shared state representation that distinguishes the states then needs at least \(N-1\) dimensions, far more than the model width. Trained models learn only a weak shared state representation and recover almost no held-out transitions (Table \ref{tab:periodic_companion}).

A second action that commutes with \(A\), even the identity, admits a shared state representation. We replace \(B\) by the identity or by an action commuting with \(A\), keeping \(A\) and the held-out transitions fixed, and train transformers and linear models compositionally at \(T=4\). Neither \(B = 1\) nor \(AB = BA\) entails the held-out transitions, which still require \(A^4 = 1\). Both models learn a stronger shared state representation and propagate the periodicity constraint (Table \ref{tab:periodic_companion}).

Forcing a shared state representation for \(A\) recovers the held-out transitions even with an arbitrary \(B\). In trained linear models, replacing \(x_{s_3}\) by \(M_A^3 x_{s_0}\) and reading out each output as the state with the nearest input vector recovers all held-out transitions. In transformers, freezing the input and output vectors to an arrangement built from seen transitions, where \(A\) acts as one matrix \(R\) with \(R^4 = I\), recovers 95\% of held-out transitions.

The inverse constraint fails in the same way when its held-out transitions are reached only through an unconstrained action. We add an arbitrary third action to the inverse world and train transformers compositionally at \(T=2\). They recover 71 to 86\% of the held-out transitions \((s, R, t)\) that are not adjacent to another. If we also hold out the transitions by \(L\) out of \(s\) and into \(t\), so that only the third action leaves \(s\) and enters \(t\), they recover none. When the third action is the identity, they recover 94 to 100\%.

\begin{table}[h]
\caption{\textbf{Commuting second actions improve the shared state representation and periodicity propagation.} Models trained compositionally at \(T=4\) on \(A^4=1\) worlds with \(N=1024\). Shared is the cosine between \(M_A x_u\) and \(x_{A(u)}\) over seen transitions; for the transformer, \(M_A\) is one matrix fitted to the state embeddings on 80\% of them and scored on the rest. Averages over three seeds.}
\label{tab:periodic_companion}
\begin{center}
\small
\begin{tabular}{lrrrr}
\toprule
 & \multicolumn{2}{c}{Transformer} & \multicolumn{2}{c}{Linear model} \\
\cmidrule(lr){2-3}\cmidrule(l){4-5}
Second action \(B\) & Shared & Held-out (\%) & Shared & Held-out (\%) \\
\midrule
Arbitrary & 0.73 & 0.0 & 0.75 & 0.3 \\
Identity & 0.96 & 99.3 & 0.86 & 44.8 \\
Commuting & 0.95 & 96.2 & 0.87 & 63.9 \\
\bottomrule
\end{tabular}
\end{center}
\end{table}

\section{Constraint propagation in planning and control}
\label{appx:control}

We test whether world models use propagated constraints to reach goals through held-out transitions. The monoid worlds are the inverse and commutative worlds of Section \ref{sec:monoid_world}. In AI2-THOR and Habitat, a random policy explores one room or home in episodes of \(20\) actions. We hold out transitions entailed by seen transitions and the constraints of Section \ref{sec:vision_world}, and never train on the rest. World models train on all \(T=4\) paths of seen transitions for 150k steps. Test goals are pairs \((s, g)\) within four actions where every shortest path from \(s\) to \(g\) crosses a held-out transition.

Search finds the shortest path to \(g\) in the world model's predicted transitions by dynamic programming. Imagination and representation train a policy \(\pi(a \mid s, g)\), an MLP with two hidden layers, on the world model's frozen embeddings of the current state and the goal. As in Dreamer \citep{hafner2020dreamer}, the imagination policy is trained by reinforcement learning only on transitions imagined by the world model, earning \(0.9^{t-1}\) for reaching \(g\) after \(t\) actions. As in World Models \citep{ha2018worldmodels}, the representation policy is trained by behavior cloning on goals whose shortest paths are fully seen. Every policy then acts in the true world for up to \(64\) actions.

\begin{figure}[h]
\centering
\includegraphics[width=\textwidth]{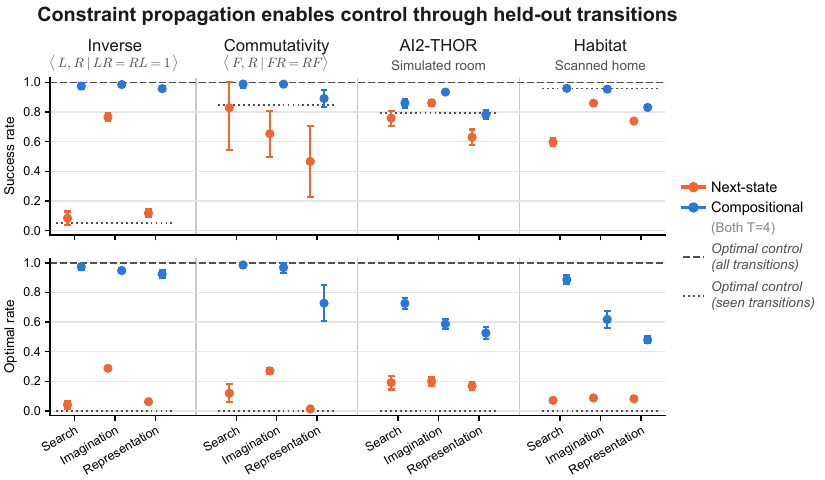}
\caption{\textbf{Compositionally trained world models reach goals by shortest paths.} Success rate (top), the fraction of test goals reached within \(64\) actions, and optimal rate (bottom), the fraction reached by a shortest path, for the setup of Figure \ref{fig:control}. Mean \(\pm\) SD over five seeds.}
\label{fig:control_rates}
\end{figure}

Seen transitions alone reach most test goals by longer paths: \(85\%\) in the commutative world, \(79\%\) in AI2-THOR, and \(96\%\) in Habitat (Figure \ref{fig:control_rates}, top). Reaching a goal by a shortest path requires propagation. Next-state world models reach at most \(29\%\) of test goals by a shortest path, against \(48\)--\(99\%\) for compositional world models (Figure \ref{fig:control_rates}, bottom).

Policies learned on compositional world models reach an average SPL of \(0.86\) with imagination and \(0.75\) with representation, against \(0.38\) and \(0.23\) on next-state world models and \(0.26\) for optimal control with seen transitions (Table \ref{tab:control_policies}).

\begin{table}[h]
\caption{\textbf{Learned policies reach held-out goals with compositionally trained world models.} SPL, mean \(\pm\) standard deviation over five seeds.}
\label{tab:control_policies}
\begin{center}
\small
\setlength{\tabcolsep}{3pt}
\begin{tabular}{llrrrrr}
\toprule
Policy & World model & Inverse & Commutative & AI2-THOR & Habitat & Mean \\
\midrule
Imagination & Next-state & 0.44 \(\pm\) 0.02 & 0.34 \(\pm\) 0.05 & 0.40 \(\pm\) 0.03 & 0.32 \(\pm\) 0.01 & 0.38 \\
 & Compositional & 0.96 \(\pm\) 0.01 & 0.97 \(\pm\) 0.03 & 0.72 \(\pm\) 0.03 & 0.78 \(\pm\) 0.05 & 0.86 \\
\midrule
Representation & Next-state & 0.08 \(\pm\) 0.02 & 0.10 \(\pm\) 0.09 & 0.38 \(\pm\) 0.04 & 0.37 \(\pm\) 0.01 & 0.23 \\
 & Compositional & 0.94 \(\pm\) 0.02 & 0.77 \(\pm\) 0.08 & 0.64 \(\pm\) 0.04 & 0.64 \(\pm\) 0.02 & 0.75 \\
\midrule
\multicolumn{2}{l}{Optimal control, seen transitions} & 0.01 \(\pm\) 0.01 & 0.14 \(\pm\) 0.09 & 0.40 \(\pm\) 0.03 & 0.50 \(\pm\) 0.01 & 0.26 \\
\bottomrule
\end{tabular}
\end{center}
\end{table}

\section{Compositional training improves world-model benchmarks}
\label{appx:literature_benchmarks}
We test whether compositional training improves world models on three published benchmarks, keeping each paper's model, data, and metric and changing only the training objective. C-SWM learns object latents from frames of three bodies under gravity and predicts the latent of the next frame \citep{kipf2020contrastive}. Compositional training predicts the latents \(T=2\) steps ahead from one frame. LeWorldModel predicts the next latent from frames and an action and is tested by planning to a goal in a two-room environment \citep{maes2026leworldmodel}. Compositional training adds prediction of the latents \(T=3\) actions ahead from one frame. WeatherBench trains a CNN to predict global pressure and temperature fields \(6\) hours ahead and forecasts \(5\) days by applying it \(20\) times \citep{rasp2020weatherbench}. Compositional training predicts the fields \(T=2\) steps ahead from the current fields. Next-state trained models reproduce the published results. Compositional training improves 10-step prediction in 3-body physics from \(74\%\) to \(90\%\), planning success from \(87\%\) to \(94\%\), and reduces the 5-day forecast error by \(42\%\). (Table \ref{tab:literature_results}).

\begin{table}[h]
\caption{\textbf{Compositional training improves world-model benchmarks.} Each paper's metric for its published model, our next-state retraining, and compositional training, in \% except weather, which is the error of \(500\) hPa geopotential in m\(^2\)s\(^{-2}\) (lower is better). Mean \(\pm\) standard deviation over \(12\) (3-body) and \(3\) (Two-Room, weather) seeds.}
\label{tab:literature_results}
\begin{center}
\small
\begin{tabular}{llrrr}
\toprule
Benchmark & Metric & Published & Next-state & Compositional \\
\midrule
3-body physics & 10-step prediction & 75.5 & 74 \(\pm\) 15 & 90 \(\pm\) 5 \\
Two-Room & Planning success & 87 & 87 \(\pm\) 1 & 94.0 \(\pm\) 0.7 \\
Weather & 3-day forecast error & 1114 & 1084 \(\pm\) 90 & 593 \(\pm\) 10 \\
 & 5-day forecast error & 1559 & 1519 \(\pm\) 89 & 874 \(\pm\) 16 \\
\bottomrule
\end{tabular}
\end{center}
\end{table}
\end{document}